\documentclass{egpubl}
\usepackage{mlvis2022}

 \WsPaper           

 \electronicVersion 

\ifpdf \usepackage[pdftex]{graphicx} \pdfcompresslevel=9
\else \usepackage[dvips]{graphicx} \fi

\PrintedOrElectronic

\usepackage{t1enc,dfadobe}

\usepackage{egweblnk}
\usepackage{cite}

\usepackage{xcolor}

\title[ViNNPruner: Visual Interactive Pruning for Deep Learning]%
      {ViNNPruner: Visual Interactive Pruning for Deep Learning}

\author[U. Schlegel \& S. Schiegg \& D. Keim]
{\parbox{\textwidth}{\centering U. Schlegel$^{1}$\orcid{0000-0002-8266-0162}, S. Schiegg$^{1}$, and
D. A. Keim$^{1}$
        }
        \\
{\parbox{\textwidth}{\centering $^1$University of Konstanz, Germany\\
      }
}
}

\begin{document}


\maketitle
\begin{abstract}
Neural networks grow vastly in size to tackle more sophisticated tasks.
In many cases, such large networks are not deployable on particular hardware and need to be reduced in size.
Pruning techniques help to shrink deep neural networks to smaller sizes by only decreasing their performance as little as possible.
However, such pruning algorithms are often hard to understand by applying them and do not include domain knowledge which can potentially be bad for user goals.
We propose ViNNPruner, a visual interactive pruning application that implements state-of-the-art pruning algorithms and the option for users to do manual pruning based on their knowledge.
We show how the application facilitates gaining insights into automatic pruning algorithms and semi-automatically pruning oversized networks to make them more efficient using interactive visualizations.


\end{abstract}  
\section{Introduction}

Deep neural networks are often considered black boxes with only limited possibilities to look into and comprehend their decision-making processes.
However, due to their state-of-the-art performance in an increasing number of tasks and the advantage of applying them to problems without domain knowledge, the necessity for explainability grows faster and faster.
Furthermore, debugging such complex models is a tedious task for developers because finding a good architecture fit for underlying data is often a trial-and-error path.
While large models such as AlexNet~\cite{krizhevsky2012imagenet} or GPT-2~\cite{radford2019language} achieve peak performances in image classification and natural language processing, such deep neural networks are not deployable in many real-world cases as these are too large for used hardware in, e.g., embedded system~\cite{frankle2018lottery}.

Neural network pruning helps shrink large-scale networks in memory and speed performance by removing neurons from a neural network model.
Automatic approaches such as MAP~\cite{han2015learning} or LAP~\cite{park2020lookahead} are well based on theoretical ground, and empiric evidence but are still not easily understandable applied to complex models.
Explainable AI (XAI) supports users in investigating their models to enable explainable decisions of such complex models~\cite{spinner2019explainer}.
Thus, using pruning methods to reduce the number of hidden neurons with visualizations to support understanding enables explainability and debugging.
CNNPruner~\cite{li2020cnnpruner} introduces a pruning application similar to the proposed in this work but focuses purely on convolutional neural networks (CNN), while ViNNPruner extends to every layer type.
The focus shifts from a general interactive XAI perspective on all kinds of neural networks onto images and CNNs with fixed pruning strategies without the option to support these with manual pruning.

In the following, we present ViNNPruner, an application for the interactive visual pruning for deep learning models (CNNs, RNNs, FCNNs, MLPs, ...) incorporating automatic pruning approaches such as LAP~\cite{park2020lookahead} as well as interactive manual pruning using neuron masks.
Visualizations of such neuron pruning masks of automatic and manual, together with feature map visualizations, enable further insights into the models' inner decision-making and pruning algorithms.
Thus, enabling model developers to use automatic and interactive approaches to incorporate domain knowledge into pruning steps.
We show findings of automatic pruning algorithms against each other and on oversized networks to highlight the options of the whole application towards pruning.

Thus, our contributions are:
(1) ViNNPruner, an application for visual interactive pruning based on identified goals and requirements,
(2) Insights into prominent automatic pruning strategies integrated into ViNNPruner,
(3) Interactive visualization techniques to support manual pruning on top of strategies to tightly integrate users with domain knowledge into pruning steps.

We interchangeably call a trained (deep) neural network either model or network in the following.
We first identify the goals and requirements we used to build the ViNNPruner components. 
Next, we describe the application focusing on components and interaction options.
Last, we show findings.
A running demo can be tested at \url{https://interactive-pruning.dbvis.de/}.

\begin{figure*}[h!t]
  \centering
  \includegraphics[trim={0cm 3.8cm 4.7cm 0cm},clip,width=1.0\linewidth]{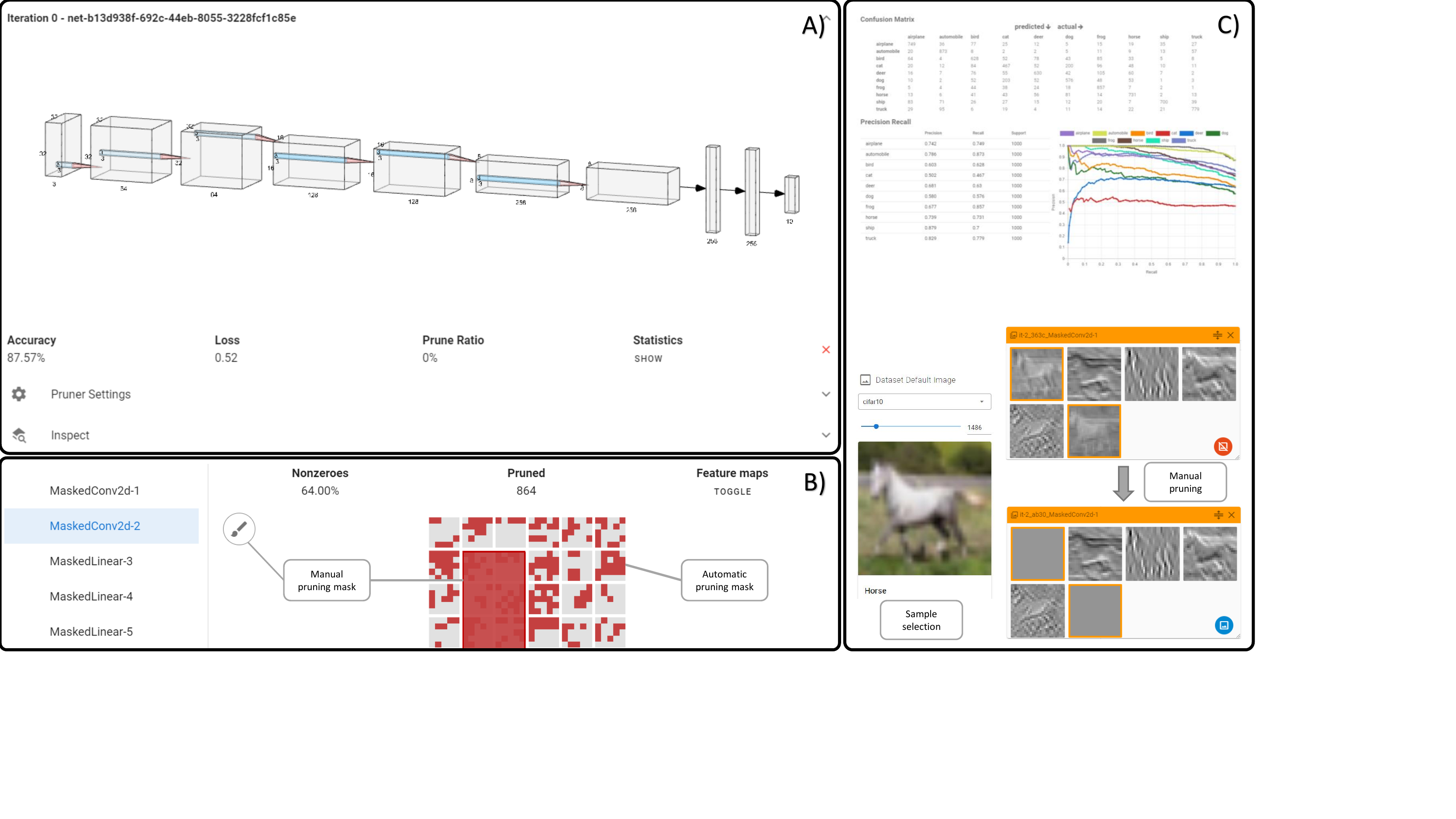}
  \caption{Overview of the three components of ViNNPruner on Cifar10 with AlexNet~\cite{krizhevsky2012imagenet}. \textbf{A)} showing the plain card view with neural network visualization and general metrics about the pruning and the performance of the network. \textbf{B)} Neural network pruning mask in red with an automatic pruning algorithm (smaller red pixels) and manual rectangle pruning (larger red rectangle) applied. \textbf{C)} On top, precision and recall metrics for the applied model and dataset with a confusion matrix and precision-recall curve. These are calculated at every step. Bottom, a selected horse from the dataset with the corresponding feature maps when the network is applied before and after pruning.}
  \label{fig:main1}
  \vspace{-1em}
\end{figure*}

\section{Related Work}

Interactive pruning of deep neural networks is tightly connected to two key research areas: state-of-the-art pruning algorithms and network layer visualizations.
Magnitude-based pruning approaches prove to be especially suitable for interactivity, as the tunable hyperparameters on a per-layer basis are quickly and effectively optimized. 
Magnitude Approach Pruning (MAP)~\cite{han2015learning} identifies prunable weights based on their significance relative to others. 
Their performance evaluation shows that removing smaller weights improves computation times by several orders of magnitude while essentially preserving network accuracy. 
The most influential hyperparameter is the percentage of removed weights for each layer.
To make better use of magnitude-pruning, Look-Ahead Pruning (LAP)~\cite{park2020lookahead} includes information from adjacent layers to determine unnecessary weights. 
Park et al.~\cite{park2020lookahead} use both forward and backward exploration for each layer to outperform MAP. 
Comparing these pruning algorithms in an interactive setting can lead to insights into the network and data as both accuracy improvements and computation performance are observable.

Forms of visualizing feature maps and weights for neural networks increase interpretability as these demonstrate fundamental insights into the model's behavior. 
CNNVis~\cite{liu_towards_2016} uses feature map visualizations alongside neuron clustering to identify critical paths in the network.
Contrary to pruning mask visualizations, such an approach emphasizes only the most activated sections of the network.
Liu et al.~\cite{liu_towards_2016} approach XAI to reduce visual noise and clutter in the visualization to facilitate identifying similar neurons in clusters.
A matrix visualization of bundled neurons facilitates assigning meaning to bundled neurons in a layer graph by reordering activated neurons. 
Correct-ordering neuron activation proves helpful for visualizing pruning masks in context by reducing noise in identifying pruning patterns.

Dividing multiple visualizations into different sections also assists in retaining an overview of large neural networks.
Summit~\cite{hohman_summit:_2019} uses compartments to allow practical inspection of deep neural networks.
Having components that provide multiple views into a neural network, like their attribution graph and embedding view, facilitates understanding of the possible purpose of each layer. 
Scalability is an essential factor, as larger networks often become very cumbersome to inspect manually. 
Hohman et al.~\cite{hohman_summit:_2019} assign categories to activated neurons that help understand which class is connected to them. 
Such an approach reduces clutter and allows for higher-level inspection of feature maps while retaining understandability.

CNNPruner~\cite{li2020cnnpruner} is an application to interactively prune convolutional neural networks (CNNs) in multiple iteration steps. 
Opposed to ViNNPruner, CNNPruner focuses actively on taking control of the pruning process with Taylor Expansion.
ViNNPruner focuses on optimizing and explaining pruning masks to facilitate the pruning process. 
However, feature map visualizations for understanding their functionality also play a role in discovering optimal pruning. 
Thus, such visualizations emphasize the importance in combination with explaining feature maps as a common denominator in different XAI tools for CNNs.
Providing visual assistance in explaining the purpose of filters is transferable to neural network pruning, thus opening the exploration of many established visualization techniques for neural network pruning.




\section{Goals and requirements to consider}
Based on related work, we see that automatic approaches of neural network pruning can lead to generally smaller and faster networks but lose prediction performance and neglect domain knowledge.
Thus, we argue for an interactive approach incorporating model experts to facilitate understanding and improving pruning.

\textcolor{darkgray}{\textbf{Goals.}}
Our user goals are based on colleagues' experiences working with neural networks in the image, text, and time series domains.
At first, an interactive pruning application needs to integrate the state-of-the-art pruning neural networks we identified in the related work section.
Through such a goal, we guarantee that even the automatic approach yields state-of-the-art results.
Next, users need to be able to deeply inspect the differences of the network over the iterations of the pruning algorithms to enable an understanding of the pruning algorithms strategy.
At last, based on domain knowledge, users need to be able to iteratively and interactively prune their network.
Thus, these goals enable an in-depth inspection of the pruning mechanism automatically and manually.


\textcolor{darkgray}{\textbf{Requirements.}}
We set some requirements for our interactive pruning application based on these goals.
The application approach needs to implement state-of-the-art pruning algorithms to fulfill the first goal.
Through our related work, we identified these as MAP~\cite{han2015learning} and LAP~\cite{park2020lookahead} as these are available (open-source implantation) and recent.
To present results and pruning actions, the application approach needs visualizations to support neural networks' presentation and the pruning algorithms' changes.
We focus on a general concept of neural network visualizations as graphs with NN-SVG~\cite{lenail2019nn} and dense pixel visualizations for the layers internally in the network.
As the last requirement, the application needs to be highly interactive, and especially the visualizations need to integrate user actions and changes to support the manual pruning and the in-depth inspection as neural networks can get quite large in memory size.
To reduce such visual clutter as much as possible and allow the identification of patterns, we segregate different steps of the pruning process to allow better comparison and influence in the pruning process.


\section{The ViNNPruner application}


ViNNPruner is based on various components supporting experts to iteratively and interactively prune neural networks incorporating the previous goals and requirements.
We combine multiple views into the network to create a workflow based on refinement and emphasis on detecting pruning patterns as they appear.
ViNNPruner loads upon startup every trained model (CNNs, RNNs, ...) available with a dataset, either the training or a test set in a specific directory.

\textcolor{darkgray}{\textbf{Components.}} 
When loading a model, ViNNPruner presents users with a network overview that features a neural network visualization using NN-SVG~\cite{lenail2019nn}, accuracy, loss metrics, and current pruning ratio. 
The card view component (Fig.~\ref{fig:main1} A)) enables quickly determining the pruning state of the network by having the most critical performance metrics available. 
Increasing the pruning ratio will scale the layer sizes accordingly to approximate how many neurons are pruned.
Each pruning step is modeled in such a minimizable card component underneath each other to show a timeline of the pruning.
These cards enable backtracking and replicability to increase the understanding of the pruning process. 
Users can remove cards if the pruning step is unsuccessful or not valid. 
All card components can be hidden or opened multiple times to compare all metrics across pruning iterations quickly.
To steer the hyperparameters of the magnitude-based pruning algorithms, the application offers a settings section in which users can select custom pruning ratios per layer to focus, e.g., on decreasing linear layers. 
To facilitate a quick start for comparing the picked algorithms, we suggest Park et al.~\cite{park2020lookahead} parameters by default.

For a more advanced inspection of a pruning step, ViNNPruner visualizes pruning masks for each layer.
These masks can be accessed by picking the desired layer in the inspect menu on the left. 
Fig.~\ref{fig:main1} B) depicts the extended view, including layer-specific metrics like the pruned and remaining number of non-zero weights.
Since the pruning mask is structurally identical to the layer definition, each row represents a channel in the currently viewed layer, with each square showing one kernel. 
Adhering to this structure permits users to quickly recognize patterns visible in channels and facilitates observing results when experimenting with channel pruning. 
Zeroed weights are colored red.
To complement the pruning mask, feature maps are also provided with a visualization for each layer.
Fig.~\ref{fig:main1} C) shows an example of how pruning can eliminate channels from the network, which would show up as a red row in the mask visualization.
As pruning can effect changes in network performance to various degrees, advanced metrics such as a confusion matrix paired with precision-recall curves allow for a fine-grained comparison between pruning steps. 
These metrics and visualizations are helpful to determine inter-class variations when eliminating specific channels or unused connections in the network.
The application further enables to select different images of the dataset to use as an input for the feature map visualizations.

\textcolor{darkgray}{\textbf{Interactions.}} 
When pruning a network, users first need to configure desired values in the settings tab. 
Specific layers can be further pruned with the advanced per-layer settings while leaving others unmodified. 
Such an approach supports applying MAP~\cite{han2015learning} and LAP~\cite{park2020lookahead} iteratively at every step. 
The resulting pruning masks can then be inspected and modified further by applying another pruning step algorithmically or manually brushing neurons that should be removed. 
After pruning, inspecting features maps for each layer reveals changes in layer outputs originating from applying a pruning method. 
Further, clicking channels in the feature map visualization, as seen in Fig.~\ref{fig:use-case2} not only allows for observing changes in the layer outputs but enables marking channels in the pruning mask for removal in the next pruning step. 
To better analyze effects when removing layers, it is also possible to restore previously removed channels by re-selecting them. 
Such an undo option is useful when using a pruning algorithm where it is of interest to determine the impact on network performance for detected pruning patterns.
To explore larger networks and layers, users can zoom and resize the pruning masks, which allows for better fine-grained tuning and serves as a mid-level between retaining detail and usability without omitting any data points.
When reaching a pruning step that heavily degrades network performance, users can choose to return to a different step or use statistical analysis in the form of precision-recall and the confusion matrix data to determine whether the pruning is optimized or requires further fine-tuning.


\begin{figure}[h!t]
  \centering
  \includegraphics[trim={1cm 8cm 25cm 1cm},clip,width=0.79\linewidth]{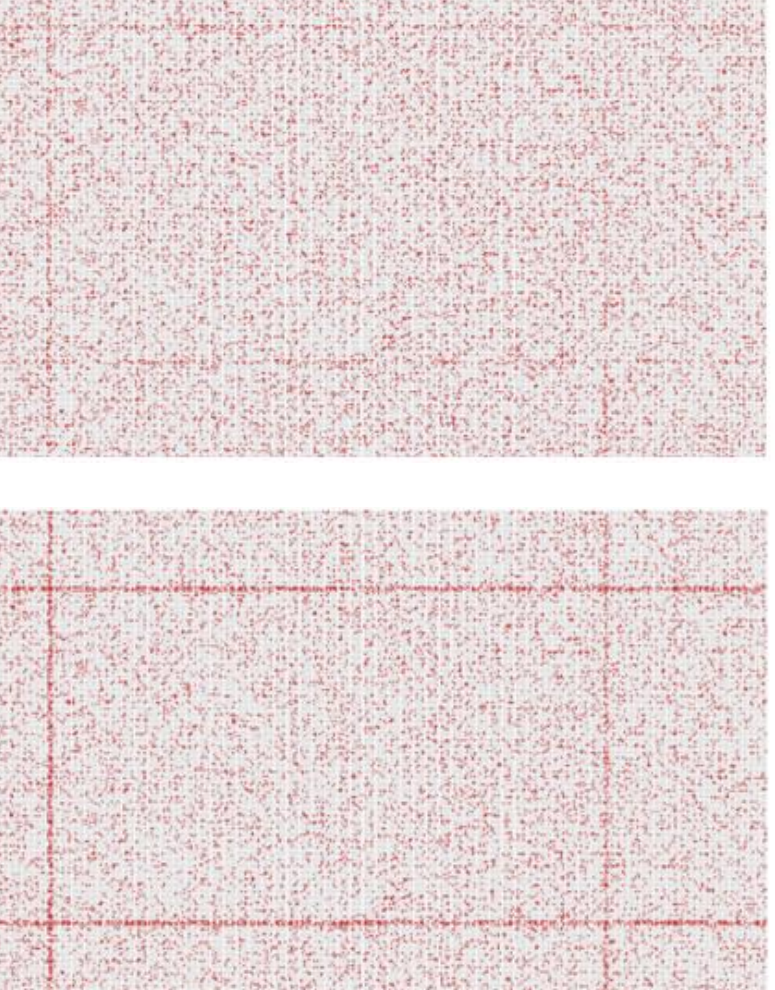}
  \caption{Use case difference MAP and LAP. MAP on top shows less patterns in the pruning mask than LAP on the bottom with a clear pattern eliminating whole features.}
  \label{fig:use-case1}
  \vspace{-2em}
\end{figure}

\begin{figure}[h!t]
  \centering
  \includegraphics[trim={10cm 6.8cm 15.8cm 1.2cm},clip,width=0.85\linewidth]{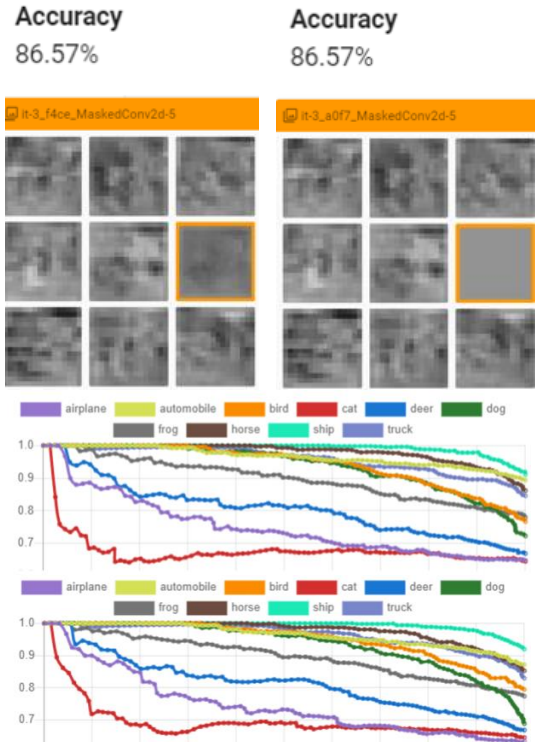}
  \caption{Use case of a whole feature map elimination without decreasing the accuracy. Similar patterns can accomplish a whole feature elimination to the bottom image of Fig.~\ref{fig:use-case1} to reduce the size of AlexNet on the Cifar10 dataset with no significant change in accuracy or precision-recall curve.}
  \label{fig:use-case2}
  \vspace{-2em}
\end{figure}

\section{Use case examples}
As we just presented how ViNNPruner tackles our goals, we further show how the application can be used to solve other tasks.

Fig.~\ref{fig:use-case1} shows the difference between our two state-of-the-art algorithms, namely MAP and LAP.
By visualizing the pruning masks for both algorithms for the same network, we see significant differences between these strategies by looking at a convolution layer in the VGG network.
MAP looks relatively uniformly to mask out different neurons on its pruning iteration.
While LAP presents some hints to eliminate whole channels in a convolution layer.
Such a finding leads to the conclusion that, in this case, in the large network, a smaller subnetwork exists, which supports the lottery ticket hypothesis~\cite{frankle2018lottery}.
The lottery ticket hypothesis assumes that there is a smaller network achieving the same prediction as the large one contained in every large network.

Pruning oversized networks is an essential task for developers and users in general as these are faster and smaller.
Further, decreasing the capacity can potentially help to have a better generalization of other data and can potentially mitigate overfitting.
Thus, ViNNPruner enables to analyze oversized networks by supporting the inspection of pruning masks and feature maps.
Many feature maps of the last convolution layer do not extract much information for the later classification layers.
Thus, these feature maps can be added to the pruning masks and eliminated for the network.
The analysis of such feature maps can be done by using the advanced view, see Fig.~\ref{fig:main1} C) bottom and Fig.~\ref{fig:use-case2}.
Such an elimination can then be done either using an automatic pruning algorithm with, e.g., LAP (Fig.~\ref{fig:use-case1}) or manual (Fig.~\ref{fig:main1} B).
In the case of AlexNet on Cifar10 and Fig.~\ref{fig:use-case2}, we can eliminate whole feature maps and do not reduce the accuracy of the model to reduce the size of our network.


\section{Conclusions and future works}
We presented ViNNPruner, an application for the interactive visual pruning of deep learning models to support automatic and manual pruning.
The application enables to compare state-of-the-art pruning techniques with their pruning masks to understand the different pruning strategies better.
Further, users can change the pruning with their domain knowledge by enabling a manual steering of these masks.
Thus, creating an environment for experts and regular users to explore deep neural network pruning.
As we showed, users can then prune oversized networks to increase speed and decrease memory usage to handle their needs better.

By further adding guidance and suggestions of potential neurons to prune, additional help for users is possible to integrate into the application.
Also, by pre-computing possible pruning settings and masks, speculative execution, similar to Sperrle et al.~\cite{sperrle2019speculative} can help to steer users into new directions and further improve interaction and performance.
Such an approach facilitates non-experts to explore more pruning steps and get a deeper understanding of these.


\bibliographystyle{eg-alpha-doi}

\bibliography{egbibsample}


\end{document}